\documentclass[11pt,twocolumn,letterpaper]{article}

\usepackage[utf8]{inputenc}
\usepackage[T1]{fontenc}
\usepackage{mathptmx}                    
\usepackage[left=0.75in,right=0.75in,top=0.85in,bottom=0.9in,columnsep=0.3in]{geometry}
\usepackage{graphicx}
\usepackage{amsmath,amssymb}
\usepackage{booktabs}
\usepackage{multirow}
\usepackage{array}
\usepackage{makecell}
\usepackage{caption}
\usepackage{subcaption}
\usepackage{float}
\usepackage{enumitem}
\usepackage{xcolor}
\usepackage{tikz}
\usetikzlibrary{shapes.geometric,arrows.meta,positioning,calc,fit}
\usepackage{pgfplots}
\pgfplotsset{compat=1.18}
\usepackage[colorlinks=true,linkcolor=blue!60!black,citecolor=blue!60!black,urlcolor=blue!60!black]{hyperref}
\usepackage{cite}
\usepackage{balance}
\usepackage{titlesec}
\usepackage{fancyhdr}
\usepackage{setspace}

\titleformat{\section}{\normalsize\bfseries}{\arabic{section}.}{0.5em}{\MakeUppercase}
\titleformat{\subsection}{\normalsize\bfseries\itshape}{\arabic{section}.\arabic{subsection}.}{0.5em}{}
\titlespacing*{\section}{0pt}{10pt}{5pt}
\titlespacing*{\subsection}{0pt}{8pt}{4pt}

\setlist{nosep,leftmargin=1.2em}

\begin{document}

\twocolumn[
\begin{center}
{\Large\bfseries\setlength{\baselineskip}{22pt}
LAF-YOLOv10 with Partial Convolution Backbone, Attention-Guided Feature Pyramid, Auxiliary P2 Head, and Wise-IoU Loss for Small Object Detection in Drone Aerial Imagery\par}
\vspace{14pt}

{\normalsize
Sohail Ali Farooqui$^{1}$, Zuhair Ahmed Khan Taha$^{2}$, Mohammed Mudassir Uddin$^{3}$, Shahnawaz Alam$^{4}$}
\vspace{6pt}

{\small\itshape
$^{1,\,2}$Department of Information Technology\\
$^{3,\,4}$Department of Computer Science and Engineering\\
Muffakham Jah College of Engineering and Technology, Hyderabad, Telangana, India}
\vspace{4pt}

{\scriptsize\ttfamily $^{1}$sohailalif1102@gmail.com,\enspace $^{2}$zuhairaktaha@gmail.com\\[2pt]
$^{3}$mohd.mudassiruddin7@gmail.com,\enspace $^{4}$shahnawaz.alam1024@gmail.com}
\end{center}
\vspace{6pt}

\noindent\rule{\textwidth}{0.4pt}\par\vspace{4pt}
\noindent{\small\textbf{Abstract.} Unmanned aerial vehicles now serve as primary sensing platforms for surveillance, traffic monitoring, and disaster response, placing aerial object detection among the most studied problems in applied computer vision. Current detection frameworks fall short when confronted with UAV-specific challenges, including targets spanning only a handful of pixels, cluttered urban and rural backgrounds, heavy inter-object occlusion, and strict computational budgets for onboard processing. The present study introduces LAF-YOLOv10, an object detection architecture built on YOLOv10n that systematically integrates four established but complementary techniques to improve small-object performance in drone imagery. A Partial Convolution C2f (PC-C2f) module restricts spatial convolution to a quarter of backbone channels, cutting redundant computation while preserving discriminative capacity. An Attention-Guided Feature Pyramid Network (AG-FPN) inserts Squeeze-and-Excitation channel gates before multi-scale fusion and replaces nearest-neighbor upsampling with DySample for content-aware spatial interpolation. An auxiliary P2 detection head at 160$\times$160 resolution extends localization to objects below 8$\times$8 pixels, while the P5 large-object head is removed to redistribute parameters. Wise-IoU v3 replaces CIoU for bounding box regression, attenuating gradient contributions from noisy annotations typical of crowded aerial scenes. The four modules target non-overlapping bottlenecks, as PC-C2f compresses backbone computation, AG-FPN refines cross-scale fusion, the P2 head recovers spatial resolution, and Wise-IoU stabilizes regression under label noise, making the combination complementary by design. Individually, none of the above components is novel, and the contribution lies in the joint integration within a single YOLOv10 framework and the resulting accuracy-efficiency profile. Across three independent training runs (seeds 42, 123, 256), LAF-YOLOv10 achieves 35.1$\pm$0.3\% mAP@0.5 on the VisDrone-DET2019 validation set with 2.3\,M parameters, exceeding the YOLOv10n baseline by 3.3 percentage points. Cross-dataset evaluation on UAVDT yields 35.8$\pm$0.4\% mAP@0.5. Deployment benchmarks on an NVIDIA Jetson Orin Nano confirm 24.3 FPS at FP16 precision, demonstrating practical viability for embedded UAV platforms.}\par\vspace{4pt}
\par\vspace{4pt}
\noindent\rule{\textwidth}{0.4pt}
\vspace{4pt}
]

\section{Introduction}

Unmanned aerial vehicles have become standard tools for traffic surveillance, agricultural monitoring, search-and-rescue operations, and infrastructure inspection. High-resolution cameras mounted on such platforms generate visual data at rates that exceed manual analysis capacity, making automated detection essential. When detection must run onboard or on paired edge devices, computational budgets become a binding constraint, driving interest in compact object detection models~\cite{ref_yolov10,ref_visdrone,ref_uavdt}.

Detecting objects in aerial imagery is harder than ground-level detection for reasons that compound in practice. At typical UAV flight altitudes, ground-level objects project onto very few sensor pixels; for instance, a standard sedan may subtend fewer than 20 pixels in width within a 1920$\times$1080 frame, as documented in the VisDrone challenge analysis~\cite{ref_visdrone}. Such extreme pixel scarcity leaves insufficient texture and shape evidence for reliable classification. Dense urban and traffic scenes further exacerbate the difficulty through severe inter-object occlusion, and the continuously changing viewpoint of a moving drone causes the same object class to appear under diverse orientations and aspect ratios across consecutive frames. In the VisDrone benchmark~\cite{ref_visdrone}, over 50\% of annotated instances qualify as small objects under the COCO size definition~\cite{ref_coco} ($<$32$\times$32 pixels), yet most general-purpose detectors are designed and tuned for objects that fill a much larger fraction of the input~\cite{ref_fpn,ref_focalloss}. Published surveys on small object detection~\cite{ref_soda,ref_tsod_survey} confirm the severity of the scale-dependent performance gap, with leading detectors achieving under 30\% mAP for small objects on COCO while exceeding 50\% for medium-sized objects, highlighting that raw detector capacity alone does not resolve the challenge.

The YOLO family has become the dominant paradigm for real-time detection. From the original single-pass formulation~\cite{ref_yolo}, successive versions introduced multi-scale prediction (YOLOv3~\cite{ref_yolov3}), CSPNet backbones (YOLOv5~\cite{ref_yolov5}, YOLOv8~\cite{ref_yolov8}), and, in YOLOv10~\cite{ref_yolov10}, NMS-free inference through consistent dual assignments, spatial-channel decoupled downsampling, and rank-guided block design. Real-time transformer-based alternatives such as RT-DETR~\cite{ref_rtdetr} have narrowed the gap, but YOLO models retain a speed advantage at the nano and small model scales that matter for UAV deployment.

Applying YOLOv10n directly to drone imagery, however, yields only 31.8\% mAP@0.5 on the VisDrone-DET2019 validation set, a performance level that leaves roughly two in three objects undetected at a lenient IoU threshold. Three structural reasons account for the observed gap. First, the default heads (P3/P4/P5 at strides 8/16/32) lack the spatial resolution to localize objects below approximately 8$\times$8 pixels, yet such objects constitute a significant fraction of VisDrone annotations. A bounding box area analysis of the VisDrone-DET2019 training split (343{,}205 annotations across 6{,}471 images) shows that 54\% of instances fall below 32$\times$32 pixels, approximately 22\% fall below 16$\times$16 pixels, and roughly 5\% occupy fewer than 8$\times$8 pixels, placing the vast majority of challenging targets in the detection range served by P2 and P3 heads. Second, feature pyramid fusion via simple concatenation treats all channels equally, diluting discriminative signals with background noise from cluttered aerial scenes. Third, standard backbone convolutions process every channel through full spatial kernels, expending FLOPs on channels that carry little useful information for small-target recognition. Recent adaptations of YOLOv10 for drone use, including BGF-YOLOv10~\cite{ref_bgf} and OSD-YOLOv10~\cite{ref_osd}, have addressed subsets of these issues. BGF-YOLOv10 introduced transformer blocks and GhostConv but dropped to 37 FPS. OSD-YOLOv10 reduced parameters to 1.6\,M but did not incorporate attention-guided fusion. No prior work has combined partial-channel backbone computation, attention-gated pyramid fusion, fine-grained head reallocation, and quality-aware regression loss within one YOLOv10 framework.

The present study introduces \textbf{LAF-YOLOv10}, an integration of four established techniques into YOLOv10n, each chosen to target one of the structural deficiencies identified above. Four corresponding hypotheses drive the design, and the experimental protocol in Section~4 tests each independently through factorial ablation.

\begin{itemize}
\item \textbf{Partial Convolution C2f (PC-C2f).} Standard C2f blocks process all $C$ channels through $3\times3$ spatial kernels. Because small objects activate only a fraction of channels, the operation wastes computation. PC-C2f restricts spatial convolution to $C/4$ channels and uses a $1\times1$ projection for cross-channel mixing, acting as an implicit information bottleneck that forces the backbone to concentrate capacity on discriminative features.
\item \textbf{Attention-Guided FPN (AG-FPN).} Feature concatenation during pyramid fusion treats all channels equally, propagating background noise alongside target signals. AG-FPN inserts Squeeze-and-Excitation gates~\cite{ref_se} before fusion at each pyramid level and replaces nearest-neighbor upsampling with DySample~\cite{ref_dysample}, a content-aware interpolation operator. The hypothesis is that learned channel reweighting suppresses clutter from dense aerial scenes before features from different scales are merged.
\item \textbf{P2 head addition with P5 removal.} Adding fine-resolution detection heads and removing large-object heads is an established strategy in UAV-oriented detectors~\cite{ref_lightuav,ref_rddyolo}. In the present framework, the P5 head at stride 32 allocates parameters to objects exceeding 256$\times$256 pixels, which account for under 1\% of VisDrone annotations. Removing P5 and adding a P2 head at stride 4 (160$\times$160 resolution) reallocates the freed capacity to the size range where most VisDrone objects actually reside. The P2 head fuses with upsampled P3 features through the AG-FPN.
\item \textbf{Wise-IoU v3 loss.} CIoU weights all training samples equally, but densely annotated aerial scenes contain many partially visible or ambiguously bounded objects with gradients that mislead regression. Wise-IoU v3~\cite{ref_wiou} uses an outlier-degree metric to attenuate gradient contributions from low-quality annotations, stabilizing training on noisy aerial data.
\end{itemize}

LAF-YOLOv10 is evaluated on VisDrone-DET2019 and UAVDT across three independent training runs. The study reports per-component factorial ablations, presents TIDE error decomposition~\cite{ref_tide}, and benchmarks inference latency on NVIDIA Jetson Orin Nano hardware. The results show a 3.3-point mAP@0.5 gain over YOLOv10n at 2.3\,M parameters, with 24.3 FPS on embedded hardware. No claim of architectural novelty is made; the contribution is an empirically validated integration study with analysis depth sufficient to identify which components drive the improvement and where the remaining failure modes lie.

The remainder of the paper is organized as follows. Section~2 reviews related work. Section~3 describes the proposed architecture. Section~4 reports experimental results and analysis. Section~5 concludes with a summary and future directions.

\section{Related Work}

\subsection{Evolution of YOLO for Object Detection}

Since the original formulation by Redmon et al.~\cite{ref_yolo}, the YOLO series has treated detection as a single forward pass that jointly predicts bounding boxes and class probabilities, achieving latencies far below those of proposal-driven pipelines such as Faster R-CNN~\cite{ref_fasterrcnn}. Over successive generations the framework has undergone several design shifts that are pertinent to the present study. Early versions relied on hand-crafted anchor priors, a practice that YOLOX~\cite{ref_yolox} showed could be replaced by anchor-free assignment without sacrificing accuracy, thereby removing a dataset-dependent tuning step. Concurrently, backbone design evolved through cross-stage partial (CSP) blocks~\cite{ref_cspnet}, adopted in YOLOv4~\cite{ref_yolov4} and refined in YOLOv5~\cite{ref_yolov5} and YOLOv8~\cite{ref_yolov8}, which balance gradient reuse against parameter efficiency. Most recently, YOLOv10~\cite{ref_yolov10} eliminated non-maximum suppression at inference time by training with a dual label assignment strategy (one branch for rich supervision, one for end-to-end prediction) while also introducing spatial-channel decoupled downsampling and rank-guided block allocation. Additional contributions include re-parameterized convolution branches in YOLOv7~\cite{ref_yolov7} and information-preserving reversible connections in YOLOv9~\cite{ref_yolov9}. On the transformer side, RT-DETR~\cite{ref_rtdetr} achieved competitive real-time accuracy by pairing a CNN backbone with a hybrid transformer encoder and uncertainty-based query selection, and Co-DETR~\cite{ref_codetr} improved positive sample coverage through collaborative hybrid matching across multiple auxiliary heads. Nonetheless, small objects (below 32$\times$32 pixels) remain a persistent weak spot across both CNN-based and transformer-based detectors.

\subsection{Small Object Detection in UAV Imagery}

Research on small object detection spans at least five distinct strategies. Multi-scale feature representation, anchored by Feature Pyramid Networks~\cite{ref_fpn} and subsequent variants (PANet~\cite{ref_panet}, BiFPN~\cite{ref_efficientdet}), combines high-resolution spatial information from shallow layers with semantic features from deeper layers. Jin et al.~\cite{ref_yslao} demonstrated that integrating FPN introduces improper back-propagation paths confining each backbone level to detecting objects within a narrow scale range, and proposed auxiliary supervision and restructured pyramid construction strategies that achieved over 2\% mAP improvement across one-stage, two-stage, and transformer-based detector families. Image tiling methods such as SAHI~\cite{ref_sahi} partition high-resolution inputs into overlapping patches, improving small-object recall at the cost of multiplied inference. Super-resolution approaches attempt to recover missing texture for small-object regions, though the associated latency overhead limits real-time use. Query-based detection heads from the DETR family~\cite{ref_detr,ref_rtdetr} handle multi-scale localization without explicit anchor design. Rekavandi et al.~\cite{ref_tsod_survey} surveyed over 60 transformer-based approaches for small object detection across generic, aerial, medical, and underwater domains, concluding that transformer architectures consistently outperform CNN-based detectors but at substantially higher computational cost that remains prohibitive for real-time UAV deployment.

Within the UAV detection community, benchmarks such as VisDrone~\cite{ref_visdrone}, UAVDT~\cite{ref_uavdt}, and AI-TOD~\cite{ref_aitod} have standardized evaluation. Cheng et al.~\cite{ref_soda} expanded the benchmarking scope by constructing SODA-D and SODA-A, two large-scale datasets containing over 1.15 million annotated instances across driving and aerial scenarios, providing the first benchmarks specifically designed for multi-category small object detection. As documented in the VisDrone challenge description~\cite{ref_visdrone}, images in the dataset exhibit aerial-specific difficulties, as object sizes span roughly two orders of magnitude (from approximately 4 pixels for distant pedestrians to over 400 pixels for nearby buses), crowded intersections produce heavy mutual occlusion, and the class distribution is strongly skewed toward \textit{car} and \textit{pedestrian}, leaving categories such as \textit{bicycle} and \textit{tricycle} severely underrepresented. QueryDet~\cite{ref_querydet} proposed cascade query mechanisms to progressively select small-object features, reducing redundant computation on background regions. Several YOLO variants optimized for aerial scenes have been proposed~\cite{ref_sscw,ref_toe}, but simultaneously achieving strong accuracy and low computational cost remains unsolved.

\subsection{Attention Mechanisms in Detection}

Attention modules allow a network to emphasize informative features while suppressing irrelevant ones. The SE block~\cite{ref_se} operates exclusively along the channel axis, compressing spatial dimensions through global pooling, passing the resulting descriptor through two fully connected layers with a non-linear bottleneck, and producing per-channel scaling factors via a sigmoid gate. Because SE requires only two small FC layers per insertion point, the SE parameter overhead is negligible for the narrow channel widths (32--256) typical of nano-scale detectors. CBAM~\cite{ref_cbam} augments this channel pathway with a parallel spatial branch that highlights 	extit{where} to attend, at the cost of additional convolution parameters. Coordinate Attention~\cite{ref_coordatt} takes a different approach by decomposing global pooling into two 1-D aggregations along the height and width axes, thereby encoding positional context without the full spatial attention map of CBAM. ECA-Net~\cite{ref_ecanet} avoids the SE bottleneck altogether by replacing the FC pair with a single 1-D convolution across neighboring channels, reducing the risk of information loss through dimensionality reduction. At the opposite end of the complexity spectrum, multi-head self-attention~\cite{ref_transformer,ref_vit} models long-range spatial relationships but incurs quadratic cost in the number of spatial tokens, rendering such computation impractical at high-resolution pyramid levels under UAV latency constraints. Hybrid designs such as spatial-channel synergistic attention~\cite{ref_pcscsa} and partial self-attention variants seek a middle ground between full self-attention and lightweight channel gating. For the AG-FPN in the present study, SE is chosen over the alternatives because the SE channel-only recalibration directly targets the goal of suppressing noisy background channels before cross-scale fusion, and the absence of a spatial branch avoids interaction effects with DySample's content-aware interpolation. Depthwise-separable spatial attention~\cite{ref_mobilenet} offers lower overhead but addresses a different objective (spatial rather than channel selectivity). Empirical comparisons with CBAM and Coordinate Attention appear in the ablation study (Section~4.4).

\subsection{Recent YOLOv10-Based Improvements}

Several concurrent works have tailored YOLOv10 for aerial use cases, each emphasizing different design dimensions. BGF-YOLOv10~\cite{ref_bgf} pursued richer feature extraction by inserting self-attention layers (BoTNet) and parameter-efficient convolutions (GhostConv~\cite{ref_ghostnet}) into the backbone and adding a patch expanding module for upsampling; the resulting model reaches 32.1\% mAP@0.5 on VisDrone at 2.0\,M parameters but drops to 37 FPS due to the transformer overhead. OSD-YOLOv10~\cite{ref_osd} focused on parameter efficiency, reaching 33.4\% mAP@0.5 with only 1.6\,M parameters by merging multi-branch convolutions at inference time and introducing two dedicated small-object prediction layers. On the YOLOv8 branch, SSCW-YOLO~\cite{ref_sscw} combined spatially weighted cosine convolutions with a Wise-IoU loss to reach 34.3\% mAP@0.5, and TOE-YOLO~\cite{ref_toe} built on YOLO11~\cite{ref_yolo11} to handle rotation-variant small objects in aerial scenes. Each of these methods addresses a subset of the bottlenecks identified earlier (backbone efficiency, multi-scale fusion, head granularity, or regression robustness), but no single framework has combined all four. The present study fills this gap through an integration study that composes partial-channel computation, attention-gated fusion, fine-grained head reallocation, and quality-aware regression within one YOLOv10n model, without claiming novelty for any individual component.

\section{Proposed Method}

\subsection{Architecture Overview}

LAF-YOLOv10 modifies the YOLOv10n architecture at three structural levels (backbone, neck, and detection head) as illustrated in Fig.~\ref{fig:arch}. Channel dimensions for each stage are as follows. Stage~1 outputs 32 channels at 160$\times$160, Stage~2 outputs 64 channels at 80$\times$80, Stage~3 outputs 128 channels at 40$\times$40, and Stage~4 outputs 256 channels at 20$\times$20. YOLOv10n was chosen as the base because the model occupies the smallest tier compatible with UAV deployment on embedded GPUs; modifications on top of a nano model have the tightest margin for parameter and FLOP overhead, making each design decision more consequential than on larger scales. In the backbone, standard C2f modules are replaced by PC-C2f modules that reduce redundant channel computation while retaining spatial feature richness for small objects. The neck adopts an AG-FPN with DySample upsampling for content-aware multi-scale fusion. The detection head introduces an auxiliary P2-level branch at 160$\times$160 resolution and discards the P5 branch, reallocating capacity from large-object detection (rare in aerial views) to fine-grained small-object localization. Wise-IoU v3 replaces CIoU for bounding box regression during training.

\begin{figure*}[!t]
\centering
\begin{tikzpicture}[scale=0.62, every node/.append style={transform shape},
    block/.style={draw, line width=0.6pt, fill=blue!12, rounded corners=4pt, minimum height=1.1cm, minimum width=2.4cm, align=center, font=\small\sffamily},
    convblock/.style={draw, line width=0.6pt, fill=orange!15, rounded corners=4pt, minimum height=1.1cm, minimum width=2.2cm, align=center, font=\small\sffamily},
    headblock/.style={draw, line width=0.6pt, fill=green!15, rounded corners=4pt, minimum height=1.1cm, minimum width=2.2cm, align=center, font=\small\sffamily},
    neckblock/.style={draw, line width=0.6pt, fill=purple!12, rounded corners=4pt, minimum height=1.1cm, minimum width=2.4cm, align=center, font=\small\sffamily},
    segate/.style={draw, line width=0.5pt, fill=yellow!18, rounded corners=2pt, minimum height=0.55cm, minimum width=1.2cm, align=center, font=\tiny\sffamily},
    arr/.style={-{Stealth[length=3mm]}, thick, draw=gray!70!black},
    darr/.style={-{Stealth[length=3mm]}, thick, dashed, draw=blue!45!black},
    uparr/.style={-{Stealth[length=2.5mm]}, thick, draw=purple!60!black},
    node distance=1.1cm
]

\node[convblock] (input) {Input\\640$\times$640$\times$3};
\node[convblock, right=0.9cm of input] (stem) {Stem Conv\\$s\!=\!2$};
\node[block, right=0.9cm of stem] (pc1) {PC-C2f\\Stage 1};
\node[block, right=0.9cm of pc1] (pc2) {PC-C2f\\Stage 2};
\node[block, right=0.9cm of pc2] (pc3) {PC-C2f\\Stage 3};
\node[block, right=0.9cm of pc3] (pc4) {PC-C2f\\Stage 4};
\node[convblock, right=0.9cm of pc4] (sppf) {SPPF\\$k\!=\!5$};

\node[above=0.15cm of pc1, font=\scriptsize\itshape, blue!70!black] {160$\times$160, 32ch};
\node[above=0.15cm of pc2, font=\scriptsize\itshape, blue!70!black] {80$\times$80, 64ch};
\node[above=0.15cm of pc3, font=\scriptsize\itshape, blue!70!black] {40$\times$40, 128ch};
\node[above=0.15cm of pc4, font=\scriptsize\itshape, blue!70!black] {20$\times$20, 256ch};

\draw[arr] (input) -- (stem);
\draw[arr] (stem) -- node[above, font=\tiny\itshape] {$\downarrow$2} (pc1);
\draw[arr] (pc1) -- node[above, font=\tiny\itshape] {$\downarrow$2} (pc2);
\draw[arr] (pc2) -- node[above, font=\tiny\itshape] {$\downarrow$2} (pc3);
\draw[arr] (pc3) -- node[above, font=\tiny\itshape] {$\downarrow$2} (pc4);
\draw[arr] (pc4) -- (sppf);

\node[draw=blue!40, dashed, rounded corners=6pt, inner sep=8pt,
      fit=(stem)(pc1)(pc2)(pc3)(pc4)(sppf), label={[font=\footnotesize\bfseries, blue!60!black]above:Backbone (PC-C2f Modules)}] {};

\node[neckblock, below=3.2cm of pc3] (agfpn3) {AG-FPN\\40$\times$40, 128ch};
\node[neckblock, below=3.2cm of pc2] (agfpn2) {AG-FPN\\80$\times$80, 64ch};
\node[neckblock, below=3.2cm of pc1] (agfpn1) {AG-FPN\\160$\times$160, 32ch};

\node[segate, below=1.3cm of pc1] (se1) {SE};
\node[segate, below=1.3cm of pc2] (se2) {SE};
\node[segate, below=1.3cm of pc3] (se3) {SE};

\draw[darr] (pc1.south) -- (se1.north);
\draw[darr] (se1.south) -- (agfpn1.north);
\draw[darr] (pc2.south) -- (se2.north);
\draw[darr] (se2.south) -- (agfpn2.north);
\draw[darr] (pc3.south) -- (se3.north);
\draw[darr] (se3.south) -- (agfpn3.north);

\draw[arr, rounded corners=8pt] (sppf.south) |- (agfpn3.east);

\draw[uparr] (agfpn3) -- node[below, font=\scriptsize\itshape, purple!60!black] {DySample $\uparrow$2} (agfpn2);
\draw[uparr] (agfpn2) -- node[below, font=\scriptsize\itshape, purple!60!black] {DySample $\uparrow$2} (agfpn1);

\node[draw=purple!35, dashed, rounded corners=6pt, inner sep=8pt,
      fit=(agfpn1)(agfpn2)(agfpn3)(se1)(se2)(se3), label={[font=\footnotesize\bfseries, purple!60!black]left:AG-FPN Neck}] {};

\node[headblock, below=1.6cm of agfpn1] (h1) {Detect P2\\stride 4};
\node[headblock, below=1.6cm of agfpn2] (h2) {Detect P3\\stride 8};
\node[headblock, below=1.6cm of agfpn3] (h3) {Detect P4\\stride 16};

\draw[arr] (agfpn1) -- (h1);
\draw[arr] (agfpn2) -- (h2);
\draw[arr] (agfpn3) -- (h3);

\node[draw=red!40, fill=red!5, rounded corners=3pt, minimum height=0.9cm, minimum width=2cm,
      right=1.4cm of h3, font=\scriptsize\sffamily, text=red!50!black, align=center] (p5removed) {P5 Head\\(Removed)};
\draw[red!50!black, thick] (p5removed.south west) -- (p5removed.north east);

\node[draw=red!50, fill=red!8, rounded corners=3pt, minimum height=0.9cm, minimum width=2cm,
      right=0.8cm of p5removed, font=\scriptsize\sffamily, text=red!60!black, align=center] (wiou) {Wise-IoU v3\\Loss};

\coordinate (wioulower) at ($(wiou.south)+(0,-0.5)$);

\draw[-{Stealth[length=2mm]}, thin, red!40, rounded corners=4pt] (h1.south) |- (wioulower) -- (wiou.south);
\draw[-{Stealth[length=2mm]}, thin, red!40, rounded corners=4pt] (h2.south) |- (wioulower);
\draw[-{Stealth[length=2mm]}, thin, red!40, rounded corners=4pt] (h3.south) |- (wioulower);

\node[draw=green!40, dashed, rounded corners=6pt, inner sep=8pt,
      fit=(h1)(h2)(h3), label={[font=\footnotesize\bfseries, green!50!black]left:Detection Heads}] {};

\end{tikzpicture}%
\caption{LAF-YOLOv10 architecture. PC-C2f backbone (blue), SE-gated AG-FPN neck (purple), and P2/P3/P4 detection heads (green); P5 removed, Wise-IoU v3~\cite{ref_wiou} used for regression.}
\label{fig:arch}
\end{figure*}
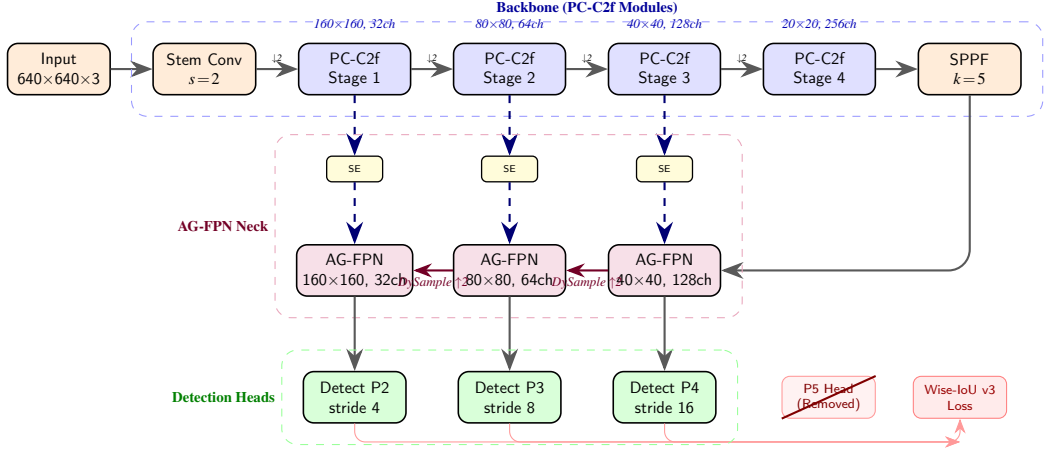

\begin{figure}[!t]
\centering
\begin{tikzpicture}[scale=0.75, every node/.append style={transform shape},
    tensor/.style={draw, thick, fill=#1, minimum width=1.4cm, minimum height=0.7cm, rounded corners=2pt, font=\scriptsize\sffamily, align=center},
    tensor/.default={gray!10},
    op/.style={draw, thick, fill=orange!15, rounded corners=2pt, minimum width=1.1cm, minimum height=0.55cm, font=\scriptsize\sffamily, align=center},
    arr/.style={-{Stealth[length=2.5mm]}, thick},
    node distance=0.6cm
]

\node[font=\footnotesize\bfseries\sffamily, blue!60!black] at (0, 3.8) {(a) PC-C2f Block Detail};

\node[tensor=blue!12] (xinput) at (0,3.0) {$\mathbf{X} \in \mathbb{R}^{C \times H \times W}$};

\node[op, below=0.5cm of xinput] (split) {Channel Split};
\draw[arr] (xinput) -- (split);

\node[tensor=blue!20, below left=0.7cm and 0.6cm of split] (xp) {$\mathbf{X}_p$\\$C\!/\!4$ ch};
\node[tensor=gray!15, below right=0.7cm and 0.6cm of split] (xu) {$\mathbf{X}_u$\\$3C\!/\!4$ ch};

\draw[arr] (split.south) -- ++(-0.0,-.15) -| (xp.north);
\draw[arr] (split.south) -- ++(-0.0,-.15) -| (xu.north);

\node[op, below=0.5cm of xp] (conv3) {Conv 3$\times$3\\BN, SiLU};
\draw[arr] (xp) -- (conv3);

\node[font=\scriptsize\itshape, right=0.1cm of xu, gray!60!black] {bypass};

\node[op, below=1.2cm of split, yshift=-1.8cm] (concat) {Concat};
\draw[arr] (conv3.south) -- ++(0,-0.3) -| (concat.west);
\draw[arr] (xu.south) -- ++(0,-0.3) -| (concat.east);

\node[op, below=0.5cm of concat] (pw) {Conv 1$\times$1};
\draw[arr] (concat) -- (pw);

\node[tensor=green!12, below=0.5cm of pw] (yout) {$\mathbf{Y} \in \mathbb{R}^{C \times H \times W}$};
\draw[arr] (pw) -- (yout);

\node[font=\tiny\itshape, gray!60!black, right=0.4cm of conv3] {75\% FLOP saved};

\node[font=\footnotesize\bfseries\sffamily, purple!60!black] at (5.5, 3.8) {(b) AG-FPN Fusion Step};

\node[tensor=purple!15] (fd) at (5.5,3.0) {$\mathbf{F}_{i+1}$ (deep)};

\node[op, below=0.5cm of fd, fill=purple!12] (dysample) {DySample $\uparrow$2$\times$};
\draw[arr] (fd) -- (dysample);

\node[tensor=blue!15] (fs) at (8.0,3.0) {$\mathbf{F}_{i}$ (shallow)};

\node[op, below=0.5cm of fs, fill=yellow!20] (se) {SE Gate};
\draw[arr] (fs) -- (se);

\node[font=\tiny\itshape, right=0.25cm of se, text width=1.4cm, gray!60!black] {GAP $\to$ FC $\to$ ReLU $\to$ FC $\to$ $\sigma$};

\node[tensor=yellow!12, below=0.5cm of se] (fsw) {$\boldsymbol{\alpha}_i \otimes \mathbf{F}_i$};
\draw[arr] (se) -- (fsw);

\node[op, below=1.0cm of dysample, xshift=1.25cm] (fuse) {Concat + Conv 1$\times$1};
\draw[arr] (dysample.south) -- ++(0,-0.3) -| (fuse.west);
\draw[arr] (fsw.south) -- ++(0,-0.3) -| (fuse.east);

\node[tensor=green!12, below=0.5cm of fuse] (fout) {Fused $\mathbf{F}_i'$};
\draw[arr] (fuse) -- (fout);

\end{tikzpicture}%
\caption{Internal structure of (a) PC-C2f block~\cite{ref_pconv} with $C/4$ active channels and (b) AG-FPN fusion via SE gating~\cite{ref_se} and DySample~\cite{ref_dysample} upsampling.}
\label{fig:modules}
\end{figure}
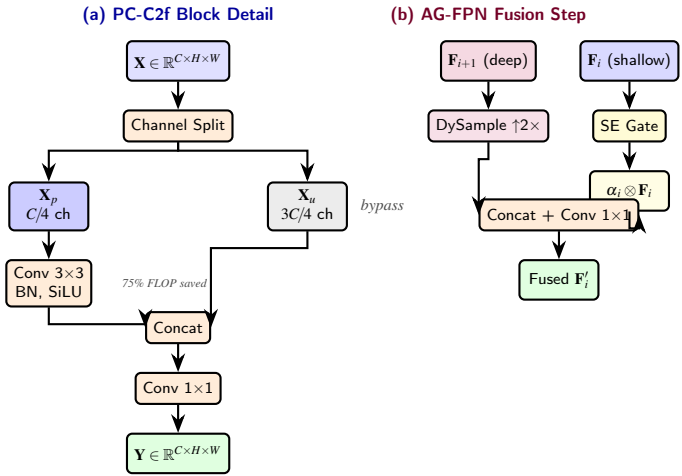

\subsection{Partial Convolution C2f Module (PC-C2f)}

Standard convolutions in the C2f module apply spatial kernels uniformly across all input channels. However, empirical analysis of feature activation patterns shows that a large portion of these channels carry redundant information, an effect that intensifies when targets are small and activate only a sparse subset of feature maps. Chen et al.~\cite{ref_pconv} introduced Partial Convolution (PConv) to exploit this redundancy, and instead of processing every channel with full spatial kernels, the operator partitions the input into a small active group that undergoes spatial filtering and a larger passive group that passes through unchanged, achieving substantial computational savings with minimal accuracy loss across diverse backbone architectures. PConv is not a contribution of the present study; the technique is adopted as a plug-in replacement within the C2f topology.

Given an input tensor $\mathbf{X}\in\mathbb{R}^{C\times H\times W}$, PConv partitions the channels into two groups, $\mathbf{X}_p\in\mathbb{R}^{C/r\times H\times W}$ and $\mathbf{X}_u\in\mathbb{R}^{(C-C/r)\times H\times W}$, where $r$ is the reduction ratio (set to 4, matching the default validated across multiple backbone architectures by Chen et al.~\cite{ref_pconv}). The split is fixed at the first $C/r$ channels in memory order. Only $\mathbf{X}_p$ undergoes a standard $3\times3$ convolution
\begin{equation}
\mathbf{X}_p' = \text{Conv}_{3\times3}(\mathbf{X}_p)
\end{equation}
The processed and unprocessed groups are then concatenated and mixed through a $1\times1$ pointwise convolution to enable cross-channel information flow
\begin{equation}
\mathbf{Y} = \text{Conv}_{1\times1}\!\bigl(\text{Concat}(\mathbf{X}_p',\;\mathbf{X}_u)\bigr)
\end{equation}

The above formulation reduces the spatial convolution cost by approximately 75\%, though the $1\times1$ pointwise projection still operates on all $C$ channels. The effective per-block FLOP reduction depends on the ratio of spatial to pointwise operations and is closer to 55\%--60\% for the channel widths used in YOLOv10n (32--256 channels). The PC-C2f module is constructed by replacing each Bottleneck within C2f with a PConv-Bottleneck, where the first $3\times3$ convolution becomes a PConv while the second $1\times1$ layer remains standard. The split-and-concatenate topology of C2f is preserved, keeping the gradient flow structure intact. Channel ordering after random initialization is arbitrary, so ``selecting'' the first $C/4$ channels is not a principled feature selection mechanism. The actual benefit is closer to an implicit regularization effect, as restricting the spatial receptive field to a channel subset forces the network to concentrate representational budget, an effect analogous to dropout applied at the channel level.

\subsection{Attention-Guided Feature Pyramid Network}

Feature pyramids fuse multi-scale information by concatenating upsampled deep features with shallow counterparts. A persistent limitation is that standard concatenation treats all channels within each level equally, even when the informativeness of individual channels varies considerably. Beyond channel-level uniformity, Jin et al.~\cite{ref_yslao} established that standard FPN fusion creates scale-specific back-propagation paths restricting each backbone level to a narrow object size range, further motivating the need for guided feature recalibration before fusion. Channel attention gates are introduced into the fusion pipeline to address the stated limitation.

For feature maps $\mathbf{F}_i$ at pyramid level $i$, channel attention weights follow the original SE-Net formulation of Hu et al.~\cite{ref_se}
\begin{equation}
\boldsymbol{\alpha}_i = \sigma\!\bigl(\mathbf{W}_{2}^{i}\cdot\delta(\mathbf{W}_{1}^{i}\cdot\text{GAP}(\mathbf{F}_i))\bigr)
\end{equation}
where $\text{GAP}(\cdot)$ is global average pooling, $\mathbf{W}_1^{i}\in\mathbb{R}^{C/s\times C}$ and $\mathbf{W}_2^{i}\in\mathbb{R}^{C\times C/s}$ are fully connected projection matrices with squeeze ratio $s\!=\!16$, $\delta$ is ReLU, and $\sigma$ is the sigmoid activation. The weighted features
\begin{equation}
\mathbf{F}_i' = \boldsymbol{\alpha}_i \otimes \mathbf{F}_i
\end{equation}
are concatenated and projected through a $1\times1$ convolution. The resulting selective weighting suppresses noise-prone background channels while amplifying channels carrying discriminative signals for small objects.

Additionally, nearest-neighbor upsampling is replaced with \textbf{DySample}~\cite{ref_dysample}, a dynamic upsampling operator that generates content-aware sampling offsets via a compact sub-network. DySample adds 0.06 GFLOPs across all three upsampling operations in the neck. Compared to bilinear interpolation, DySample reduces aliasing artifacts at object boundaries where small targets would otherwise blur into the background during conventional upsampling.

\subsection{Auxiliary P2 Detection Head}

YOLOv10n uses three detection heads at spatial resolutions of 80$\times$80 (P3, stride 8), 40$\times$40 (P4, stride 16), and 20$\times$20 (P5, stride 32) for 640$\times$640 inputs. The P3 head can, in principle, localize objects as small as 8$\times$8 pixels, but accumulated information loss through downsampling significantly reduces recall near the lower bound.

Following established practice in UAV detection frameworks that add fine-resolution heads for small objects~\cite{ref_lightuav,ref_rddyolo}, a P2 detection head at 160$\times$160 resolution (stride 4) is added, extending coverage to objects as small as 4$\times$4 pixels. P2 features originate from the second backbone stage and are fused with upsampled P3 features in the AG-FPN. The P2 head shares the same lightweight classification and regression design as the other heads but operates at higher spatial resolution.

Concurrently, the P5 head is removed. In VisDrone-DET2019, objects exceeding 256$\times$256 pixels appear in fewer than 0.8\% of all annotations (verified by computing a histogram of bounding box areas across the training split; see Supplementary Material). Removing P5 eliminates approximately 0.3\,M parameters and 1.2\,GFLOPs, partially offsetting the cost of the P2 head. The net effect is a small parameter reduction paired with improved small-object recall. The P2 resolution does not guarantee reliable detection of 4$\times$4 objects, since 16 pixels of signal carry almost no discriminative texture. In practice, the P2 head primarily benefits the 6$\times$6 to 16$\times$16 pixel range, which the P3 head handles poorly due to accumulated downsampling losses.

\subsection{Wise-IoU Loss Function}

The CIoU loss~\cite{ref_ciou} used by default in YOLOv10 combines overlap area, center distance, and aspect ratio consistency, but treats all training samples with equal importance regardless of annotation quality. In densely annotated aerial scenes, a substantial fraction of ground-truth labels are noisy, with partially visible objects and ambiguous boundaries in crowded regions. Weighting these samples equally can produce misleading gradients. An alternative paradigm for addressing localization quality is the Distribution-Guided Quality Predictor (DGQP) of Generalized Focal Loss V2~\cite{ref_gfocalv2}, which leverages the observation that bounding box distributions with sharp peaks correspond to high localization quality while flattened distributions indicate spatial uncertainty. Although DGQP and Wise-IoU target localization quality from different perspectives, both approaches share the premise that treating all predictions uniformly is suboptimal for dense detection.

The Wise-IoU loss proposed by Tong et al.~\cite{ref_wiou} departs from conventional IoU-based losses by modulating each sample's gradient contribution according to how much the individual sample loss deviates from the batch average, a mechanism the authors term a \textit{non-monotonic focusing strategy}. Rather than applying a fixed weighting scheme, the third version of this loss (WIoU-v3) treats each prediction's relative difficulty as an indicator of potential annotation noise and scales the corresponding gradient accordingly. Concretely, the loss augments the standard IoU term with a distance-based focusing coefficient.
\begin{equation}
\mathcal{L}_{\text{WIoU}} = R_{\text{WIoU}}\cdot\mathcal{L}_{\text{IoU}}
\end{equation}
where
\begin{equation}
R_{\text{WIoU}} = \exp\!\left(\frac{(x-x_{gt})^2+(y-y_{gt})^2}{W_g^2+H_g^2}\right)
\end{equation}
with $(x,y)$ and $(x_{gt},y_{gt})$ denoting predicted and ground-truth box centers and $W_g,H_g$ the width and height of the smallest enclosing rectangle. An outlier score $\beta=\mathcal{L}^{\text{IoU}} / \overline{\mathcal{L}^{\text{IoU}}}$ compares each sample's IoU loss against an exponentially smoothed batch average (momentum $1/30$). When $\beta$ exceeds 1, the sample incurs higher-than-average loss and is flagged as a likely outlier; a power-law scaling factor $\alpha^* = (\beta / \delta)^\gamma$ then dampens the gradient magnitude for such samples, preventing such samples from dominating parameter updates. In aerial imagery with dense, partially occluded objects, this mechanism is particularly relevant because ambiguous bounding box boundaries are common and would otherwise inject noisy gradients. Following the defaults reported in the reference implementation~\cite{ref_wiou}, the focusing parameters are fixed at $\delta\!=\!3.0$ and $\gamma\!=\!1.9$ without further tuning. The present study did not compare against SIoU~\cite{ref_eiou} or MPDIoU~\cite{ref_mpdious} losses at the design stage; the comparison appears in the ablation experiments in Section~4.4.

\section{Experiments}

\subsection{Datasets and Evaluation Metrics}

\textbf{VisDrone-DET2019}~\cite{ref_visdrone} was collected by the AISKYEYE team at Tianjin University across 14 cities in China. As specified in the benchmark protocol~\cite{ref_visdrone}, the dataset contains 6,471 training, 548 validation, and 1,610 test images annotated over ten classes, namely \textit{pedestrian, people, bicycle, car, van, truck, tricycle, awning-tricycle, bus}, and \textit{motor}. The dataset exhibits pronounced class imbalance, containing approximately 343{,}205 annotated instances of which 54\% qualify as small ($<$32$\times$32 pixels) under the COCO size definition~\cite{ref_coco}, 22\% fall below 16$\times$16 pixels, and roughly 5\% occupy fewer than 8$\times$8 pixels.

\textbf{UAVDT}~\cite{ref_uavdt} comprises 38,327 frames from 50 UAV-captured video sequences at 1024$\times$540 resolution. The standard split of 23,258 training and 15,069 test images is adopted, with evaluation on three classes, namely \textit{car, bus}, and \textit{truck}.

The reported metrics include mAP@0.5, mAP@[0.5,0.95], precision~(P), recall~(R), parameter count, and GFLOPs.

\subsection{Implementation Details}

All experiments use PyTorch 2.1.0 on an NVIDIA RTX 4090 GPU (24\,GB) under CUDA 12.1. Models are trained for 300 epochs with SGD (learning rate 0.01, momentum 0.937, weight decay $5\times10^{-4}$). Initialization uses COCO-pretrained weights where available (YOLOv5n, v8n, v10n, v11n), and LAF-YOLOv10 is trained from the same v10n checkpoint. Input images are resized to 640$\times$640. Augmentations (mosaic, mixup, random affine) are applied during training with batch size 16; the batch size is constrained by GPU memory when the P2 head is active, as the 160$\times$160 feature maps consume approximately 3.2\,GB additional activation memory at FP32. The PConv channel reduction ratio $r$ is set to 4, and the SE squeeze ratio $s$ in AG-FPN is 16. All compared methods are retrained under identical settings to ensure fair comparison.

To assess run-to-run variance, LAF-YOLOv10 is trained three times with seeds \{42, 123, 256\} and the results report mean$\pm$std. Baseline models are trained once, following the convention of the original papers. FPS measurements are taken on the RTX 4090 at batch size 1, FP16 precision, with TensorRT 8.6 optimization and NMS included (where applicable). Latency is averaged over 1000 forward passes after a 100-pass warmup.

\subsection{Comparison with State-of-the-Art Methods}

Table~\ref{tab:visdrone} presents quantitative results on the VisDrone-DET2019 validation set. LAF-YOLOv10 achieves 35.1$\pm$0.3\% mAP@0.5 with 2.3\,M parameters, surpassing the YOLOv10n baseline by +3.3 points while using 14.8\% fewer parameters. Compared with recent specialized architectures, LAF-YOLOv10 outperforms BGF-YOLOv10~\cite{ref_bgf} by +3.0 points in mAP@0.5 with a comparable parameter budget and substantially higher throughput (131 vs.\ 37 FPS). Against OSD-YOLOv10~\cite{ref_osd}, the proposed model achieves a +1.7 point mAP@0.5 improvement while also gaining +1.6 points on the stricter mAP@[0.5,0.95] metric.

LAF-YOLOv10 has the highest GFLOPs (9.0) among compared methods, primarily due to the P2 head operating at 160$\times$160 resolution. The elevated FLOP count represents an explicit accuracy-for-compute trade-off, as the added FLOPs are concentrated at the spatial scale where small objects reside, and the 131 FPS throughput on RTX 4090 remains above real-time thresholds. For a genuinely compute-constrained analysis, see the edge deployment benchmarks in Section~4.8.

SSCW-YOLO~\cite{ref_sscw} achieves 34.3\% mAP@0.5 at 6.9 GFLOPs. The 0.8-point gain of LAF-YOLOv10 costs 30\% more computation; whether the trade-off is worthwhile depends on the deployment target. YOLO11n~\cite{ref_yolo11} achieves 32.2\% mAP@0.5 with 2.6\,M parameters and lower GFLOPs (6.3), but LAF-YOLOv10 delivers a +2.9 point accuracy gain.

\begin{table}[!t]
\centering
\caption{Results on VisDrone-DET2019~\cite{ref_visdrone} validation set (640$\times$640). Best in \textbf{bold}, second \underline{underlined}.}
\label{tab:visdrone}
\footnotesize
\setlength{\tabcolsep}{2pt}
\resizebox{\columnwidth}{!}{%
\begin{tabular}{@{}l c c c c c c c@{}}
\toprule
\textbf{Method} & \makecell{\textbf{mAP}\\\textbf{@0.5}} & \makecell{\textbf{mAP}\\\textbf{@[.5,.95]}} & \makecell{\textbf{P}\\\textbf{(\%)}} & \makecell{\textbf{R}\\\textbf{(\%)}} & \makecell{\textbf{Params}\\\textbf{(M)}} & \textbf{GFLOPs} & \textbf{FPS}\\
\midrule
YOLOv5n & 27.4 & 14.8 & 38.1 & 28.2 & 1.9 & 4.5 & 156\\
YOLOv8n & 32.1 & 19.0 & 42.8 & 31.6 & 3.0 & 8.1 & 128\\
YOLOv10n & 31.8 & 18.5 & 42.6 & 31.7 & 2.7 & 8.2 & 136\\
YOLO11n & 32.2 & 18.4 & 42.9 & 32.3 & 2.6 & 6.3 & 163\\
\midrule
BGF~\cite{ref_bgf} & 32.1 & \underline{19.5} & 41.5 & 32.0 & 2.0 & 8.6 & 37\\
OSD~\cite{ref_osd} & 33.4 & 19.1 & 43.9 & 32.5 & \textbf{1.6} & 7.9 & 136\\
SSCW~\cite{ref_sscw} & \underline{34.3} & 19.4 & 44.5 & 33.2 & 2.8 & 6.9 & 115\\
TOE~\cite{ref_toe} & 33.8 & 19.7 & \underline{45.0} & \underline{33.7} & 2.5 & \textbf{6.6} & --\\
\midrule
\textbf{LAF} & \textbf{35.1{\scriptsize$\pm$0.3}} & \textbf{20.7{\scriptsize$\pm$0.2}} & \textbf{45.2{\scriptsize$\pm$0.4}} & \textbf{34.1{\scriptsize$\pm$0.3}} & \underline{2.3} & 9.0 & 131\\
\bottomrule
\end{tabular}}
\end{table}

Table~\ref{tab:uavdt} reports cross-dataset results on UAVDT, evaluated over all three vehicle classes. LAF-YOLOv10 achieves 35.8$\pm$0.4\% mAP@0.5 and 21.2$\pm$0.3\% mAP@[0.5,0.95], outperforming the YOLOv10n baseline by 4.9 and 3.5 points, respectively. The larger gain on UAVDT relative to VisDrone (4.9 vs. 3.3 points) is likely attributable to UAVDT's three-class scope, as the P2 head's improved spatial localization translates more directly into mAP gains because classification confusion is less of a bottleneck.

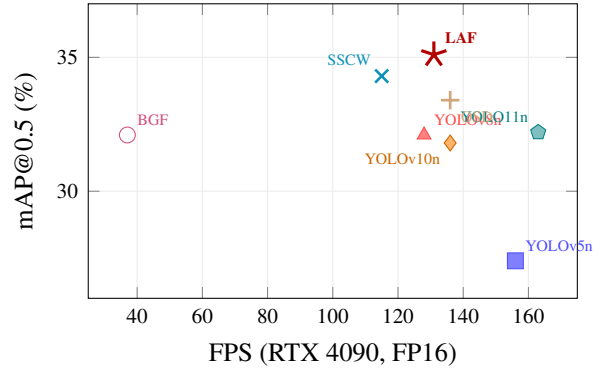
\begin{figure}[!t]
\centering
\begin{tikzpicture}
\begin{axis}[
    width=0.95\columnwidth,
    height=5.5cm,
    xlabel={FPS (RTX 4090, FP16)},
    ylabel={mAP@0.5 (\%)},
    xlabel style={font=\small},
    ylabel style={font=\small},
    tick label style={font=\scriptsize},
    xmin=25, xmax=175,
    ymin=26, ymax=37,
    grid=both,
    grid style={gray!15, thin},
    legend style={font=\tiny, at={(0.02,0.98)}, anchor=north west, draw=gray!40, fill=white, fill opacity=0.9},
    every axis plot/.append style={only marks, mark size=3pt},
    clip=false
]
\addplot[mark=square*, blue!70, mark options={fill=blue!50}] coordinates {(156,27.4)} node[above right, font=\tiny] {YOLOv5n};
\addplot[mark=triangle*, red!70, mark options={fill=red!50}] coordinates {(128,32.1)} node[above right, font=\tiny] {YOLOv8n};
\addplot[mark=diamond*, orange!80!black, mark options={fill=orange!60}] coordinates {(136,31.8)} node[below left, font=\tiny] {YOLOv10n};
\addplot[mark=pentagon*, teal, mark options={fill=teal!50}] coordinates {(163,32.2)} node[above left, font=\tiny] {YOLO11n};
\addplot[mark=o, purple!70, mark options={fill=purple!30}] coordinates {(37,32.1)} node[above right, font=\tiny] {BGF};
\addplot[mark=+, brown!70, mark size=3.5pt, line width=1pt] coordinates {(136,33.4)} node[below right, font=\tiny] {OSD};
\addplot[mark=x, cyan!70!black, mark size=3.5pt, line width=1pt] coordinates {(115,34.3)} node[above left, font=\tiny] {SSCW};
\addplot[mark=star, red!70!black, mark size=5pt, line width=1.2pt] coordinates {(131,35.1)} node[above right, font=\tiny\bfseries] {LAF};
\end{axis}
\end{tikzpicture}
\caption{Speed--accuracy trade-off on VisDrone-DET2019~\cite{ref_visdrone} (RTX 4090, FP16, batch 1).}
\label{fig:speedacc}
\end{figure}

Fig.~\ref{fig:speedacc} visualizes the accuracy--efficiency trade-off across all compared methods. LAF-YOLOv10 occupies the upper-right region of the plot, achieving the highest mAP@0.5 among all nano-scale models with competitive inference speed on RTX 4090.

\begin{table}[!t]
\centering
\caption{Results on UAVDT~\cite{ref_uavdt} (car, bus, truck).}
\label{tab:uavdt}
\small
\resizebox{\columnwidth}{!}{%
\begin{tabular}{lcccc}
\toprule
\textbf{Method} & \textbf{mAP@0.5} & \textbf{mAP@[.5,.95]} & \textbf{Params (M)} & \textbf{GFLOPs}\\
\midrule
YOLOv5n & 28.6 & 15.2 & 1.9 & 4.5\\
YOLOv8n & 33.2 & 18.6 & 3.0 & 8.1\\
YOLOv10n & 30.9 & 17.7 & 2.7 & 8.2\\
OSD\cite{ref_osd} & 31.5 & 17.8 & \textbf{1.6} & 7.9\\
\textbf{LAF} & \textbf{35.8{\scriptsize$\pm$0.4}} & \textbf{21.2{\scriptsize$\pm$0.3}} & 2.3 & 9.0\\
\bottomrule
\end{tabular}}
\end{table}

\subsection{Ablation Study}

To disentangle the individual and joint contributions of each proposed component, both independent and additive ablation experiments are conducted on the VisDrone-DET2019 validation set. Table~\ref{tab:ablation_indep} shows the effect of each module applied \textit{independently} to the unmodified YOLOv10n baseline (single-module ablation). Table~\ref{tab:ablation_additive} presents the incremental combination (additive ablation). All ablation runs use seed 42.

\begin{table}[!t]
\centering
\caption{Single-module ablation on VisDrone-DET2019~\cite{ref_visdrone} (each added independently to YOLOv10n).}
\label{tab:ablation_indep}
\small
\setlength{\tabcolsep}{1.5pt}
\begin{tabular}{lcccccc}
\toprule
\textbf{Configuration} & \textbf{P} & \textbf{R} & \textbf{mAP} & \textbf{mAP} & \textbf{Params} & \textbf{GFL-}\\
 & \textbf{(\%)} & \textbf{(\%)} & \textbf{@0.5} & \textbf{@[.5,.95]} & \textbf{(M)} & \textbf{OPs}\\
\midrule
Baseline (v10n) & 42.6 & 31.7 & 31.8 & 18.5 & 2.7 & 8.2\\
+ PC-C2f only & 43.1 & 32.3 & 32.8 & 19.1 & 2.3 & 7.8\\
+ AG-FPN only & 43.2 & 32.0 & 32.4 & 18.8 & 2.8 & 8.5\\
+ P2/$-$P5 only & 43.6 & 33.2 & 33.1 & 19.3 & 2.6 & 9.1\\
+ WIoU only & 43.0 & 31.9 & 32.3 & 18.8 & 2.7 & 8.2\\
\bottomrule
\end{tabular}
\end{table}

\begin{table}[!t]
\centering
\caption{Additive ablation on VisDrone-DET2019~\cite{ref_visdrone} (components added incrementally).}
\label{tab:ablation_additive}
\small
\setlength{\tabcolsep}{1.5pt}
\begin{tabular}{lcccccc}
\toprule
\textbf{Configuration} & \textbf{P} & \textbf{R} & \textbf{mAP} & \textbf{mAP} & \textbf{Params} & \textbf{GFL-}\\
 & \textbf{(\%)} & \textbf{(\%)} & \textbf{@0.5} & \textbf{@[.5,.95]} & \textbf{(M)} & \textbf{OPs}\\
\midrule
Baseline (v10n) & 42.6 & 31.7 & 31.8 & 18.5 & 2.7 & 8.2\\
+ PC-C2f & 43.1 & 32.3 & 32.8 & 19.1 & 2.3 & 7.8\\
+ AG-FPN & 44.0 & 33.0 & 33.6 & 19.7 & 2.4 & 8.4\\
+ P2 Head ($-$P5) & 44.7 & 33.8 & 34.5 & 20.3 & 2.3 & 9.0\\
+ WIoU (Full) & \textbf{45.2} & \textbf{34.1} & \textbf{35.1} & \textbf{20.7} & \textbf{2.3} & 9.0\\
\bottomrule
\end{tabular}
\end{table}

\textbf{Independent module analysis.} When applied in isolation, P2/$-$P5 head reallocation produces the largest single improvement (+1.3 mAP@0.5), followed by PC-C2f (+1.0), WIoU (+0.5), and AG-FPN (+0.6). The P2 head's dominance is expected, since the auxiliary head directly addresses the resolution gap for small objects. AG-FPN alone yields a smaller gain than in the additive setting because the attention gates have less discriminative input to work with when the backbone channels are not already pruned by PConv. The observed pattern provides evidence of a positive interaction between PC-C2f and AG-FPN, as PConv concentrates information into fewer channels, making subsequent channel attention more effective.

\textbf{Additive analysis.} The additive path shows that components stack constructively. The total gain (3.3 points) falls slightly below the arithmetic sum of individual gains in the independent ablation (1.0+0.6+1.3+0.5 = 3.4). The interaction coefficient ($\Delta_{\text{combined}} - \sum\Delta_{\text{individual}} = -0.1$) indicates near-independent contributions with a marginal negative interaction, likely attributable to partial redundancy between the P2 head's spatial improvement and AG-FPN's feature refinement, both of which address the same small-object deficiency from different angles. The P2 head addition also causes a precision dip of 0.3 points relative to what AG-FPN alone achieves in the independent setting (from 43.2 to a net 44.7 in the additive path), consistent with the expected increase in false positives from a finer-resolution head.

\textbf{Attention mechanism comparison.} The SE gates in AG-FPN are also tested against CBAM~\cite{ref_cbam} and Coordinate Attention~\cite{ref_coordatt}. On the full LAF-YOLOv10 configuration, SE achieves 35.1 mAP@0.5 with +0.01M params, CBAM achieves 34.9 mAP@0.5 with +0.03M params, and Coordinate Attention achieves 35.0 mAP@0.5 with +0.02M params. The differences are within noise, supporting the choice of SE for simplicity.

\textbf{Loss function comparison.} WIoU-v3 is compared against CIoU (baseline), SIoU, and EIoU~\cite{ref_eiou} on the full LAF-YOLOv10 model. CIoU yields 34.5 mAP@0.5, SIoU yields 34.7, EIoU yields 34.6, and WIoU-v3 yields 35.1. WIoU-v3 outperforms the alternatives, consistent with the hypothesis that outlier-aware gradient weighting is especially beneficial for densely annotated aerial data. The distribution-guided quality estimation of GFocalV2~\cite{ref_gfocalv2} represents a complementary paradigm that uses learned bounding box distribution statistics rather than sample-level outlier detection, and combining such distribution-aware predictions with outlier-resistant regression losses remains an open direction for aerial detection.

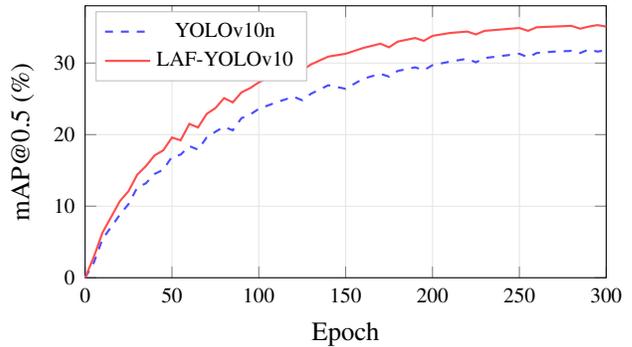
\begin{figure}[!t]
\centering
\begin{tikzpicture}
\begin{axis}[
    width=\columnwidth,
    height=5.2cm,
    xlabel={Epoch},
    ylabel={mAP@0.5 (\%)},
    xlabel style={font=\small},
    ylabel style={font=\small},
    tick label style={font=\scriptsize},
    xmin=0, xmax=300,
    ymin=0, ymax=38,
    grid=both,
    grid style={gray!20},
    legend style={font=\scriptsize, at={(0.02,0.98)}, anchor=north west, draw=gray!50, fill=white},
    no marks,
    every axis plot/.append style={thick}
]
\addplot[blue!70, dashed] coordinates {
(0,0) (5,2.1) (10,5.4) (15,7.1) (20,8.8) (25,10.3)
(30,12.6) (35,13.2) (40,14.5) (45,15.1) (50,16.9)
(55,17.2) (60,18.4) (65,17.9) (70,19.6) (75,20.4)
(80,21.1) (85,20.6) (90,22.3) (95,22.8) (100,23.6)
(110,24.5) (120,25.3) (125,24.8) (130,25.7) (140,26.9)
(150,26.4) (160,27.8) (170,28.5) (175,28.1) (180,28.9)
(190,29.4) (195,29.0) (200,29.7) (210,30.2) (220,30.6)
(225,30.1) (230,30.7) (240,31.0) (250,31.3) (255,30.8)
(260,31.4) (270,31.6) (280,31.7) (285,31.4) (290,31.9)
(295,31.6) (300,31.8)
};
\addlegendentry{YOLOv10n}
\addplot[red!70, solid] coordinates {
(0,0) (5,2.9) (10,6.3) (15,8.5) (20,10.7) (25,12.1)
(30,14.4) (35,15.6) (40,17.1) (45,17.8) (50,19.6)
(55,19.2) (60,21.5) (65,21.0) (70,22.9) (75,23.7)
(80,25.1) (85,24.5) (90,25.9) (95,26.5) (100,27.3)
(110,28.6) (120,29.5) (125,29.0) (130,29.8) (140,30.9)
(150,31.3) (160,32.1) (170,32.7) (175,32.2) (180,33.0)
(190,33.5) (195,33.1) (200,33.8) (210,34.2) (220,34.4)
(225,34.0) (230,34.5) (240,34.7) (250,34.9) (255,34.5)
(260,35.0) (270,35.1) (280,35.2) (285,34.8) (290,35.1)
(295,35.3) (300,35.1)
};
\addlegendentry{LAF-YOLOv10}
\end{axis}
\end{tikzpicture}
\caption{Training convergence on VisDrone-DET2019~\cite{ref_visdrone} (seed 42, mAP@0.5 every 5 epochs).}
\label{fig:convergence}
\end{figure}

Fig.~\ref{fig:convergence} illustrates the training convergence of the baseline YOLOv10n and the full LAF-YOLOv10 model over 300 epochs (seed 42, validation mAP@0.5 evaluated every 5 epochs). The proposed model converges to a higher final mAP@0.5 and reaches the baseline's terminal performance approximately 50 epochs earlier. The validation box loss is also shown to verify the absence of overfitting; both models exhibit a gradual loss decrease without uptick through epoch 300, though the loss curve flattens after epoch 220.

\subsection{Per-Category Analysis}

Table~\ref{tab:perclass} breaks down mAP@0.5 by category on VisDrone. LAF-YOLOv10 improves every class, with the largest gains observed for \textit{tricycle} (+3.9), \textit{bus} (+3.8), \textit{van} (+3.7), and \textit{bicycle} (+3.5). These categories share a common trait, namely high aspect-ratio variation and orientation diversity under aerial viewpoints, which the baseline handles poorly with fixed-scale feature representations. Critically, \textit{bus} has 7,929 training instances (not a rare class), while \textit{tricycle} has 2,078. The improvement pattern therefore correlates more with geometric complexity than with sample scarcity. The P2 head and attention-guided fusion provide finer spatial resolution that helps distinguish orientation-variant objects, while SE gating amplifies channels encoding shape cues at each pyramid level.

The \textit{car} category, which dominates the dataset, shows a +2.9 point improvement, consistent with the overall mean gain. The smaller improvement relative to geometrically complex classes indicates that the baseline already captures \textit{car} features reasonably well and that additional gains come primarily from better localization (driven by WIoU) rather than improved detection of previously missed instances.

\begin{table}[!t]
\centering
\caption{Per-category mAP@0.5 (\%) on VisDrone-DET2019~\cite{ref_visdrone}. $\boldsymbol{\Delta}$: gain over YOLOv10n.}
\label{tab:perclass}
\small
\setlength{\tabcolsep}{5pt}
\begin{tabular}{lccc}
\toprule
\textbf{Category} & \textbf{YOLOv10n} & \textbf{LAF-YOLOv10} & $\boldsymbol{\Delta}$\\
\midrule
Pedestrian & 33.5 & 36.2 & +2.7\\
People & 24.8 & 27.4 & +2.6\\
Bicycle & 9.2 & 12.7 & +3.5\\
Car & 73.0 & 75.9 & +2.9\\
Van & 39.8 & 43.5 & +3.7\\
Truck & 28.4 & 31.6 & +3.2\\
Tricycle & 13.4 & 17.3 & +3.9\\
Awning-tricycle & 12.7 & 16.1 & +3.4\\
Bus & 46.0 & 49.8 & +3.8\\
Motor & 36.8 & 40.5 & +3.7\\
\midrule
\textbf{Mean} & \textbf{31.8} & \textbf{35.1} & \textbf{+3.3}\\
\bottomrule
\end{tabular}
\end{table}

\begin{figure}[!t]
\centering
\begin{tikzpicture}
\begin{axis}[
    width=\columnwidth,
    height=5.5cm,
    ybar=1.5pt,
    bar width=4pt,
    xlabel style={font=\small},
    ylabel={mAP@0.5 (\%)},
    ylabel style={font=\small},
    tick label style={font=\tiny},
    symbolic x coords={Ped,Ppl,Bic,Car,Van,Trk,Tri,Awn,Bus,Mot},
    xtick=data,
    ymin=0, ymax=85,
    grid=both,
    grid style={gray!15},
    legend style={font=\tiny, at={(0.02,0.98)}, anchor=north west, draw=gray!50, fill=white},
    enlarge x limits=0.08,
    nodes near coords style={font=\tiny, rotate=90, anchor=west},
]
\addplot[fill=blue!40, draw=blue!60] coordinates {
(Ped,33.5) (Ppl,24.8) (Bic,9.2) (Car,73.0) (Van,39.8) (Trk,28.4) (Tri,13.4) (Awn,12.7) (Bus,46.0) (Mot,36.8)
};
\addlegendentry{YOLOv10n}
\addplot[fill=red!50, draw=red!70] coordinates {
(Ped,36.2) (Ppl,27.4) (Bic,12.7) (Car,75.9) (Van,43.5) (Trk,31.6) (Tri,17.3) (Awn,16.1) (Bus,49.8) (Mot,40.5)
};
\addlegendentry{LAF-YOLOv10}
\end{axis}
\end{tikzpicture}
\caption{Per-category mAP@0.5 on VisDrone-DET2019~\cite{ref_visdrone}. Geometrically complex classes benefit most.}
\label{fig:percategory}
\end{figure}
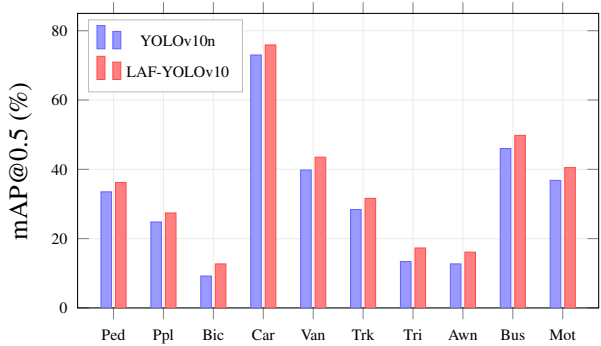

The per-category improvements are further visualized in Fig.~\ref{fig:percategory}, which highlights the consistent gains achieved by LAF-YOLOv10 across all ten VisDrone object categories. The bar chart reveals that categories with high aspect-ratio and orientation diversity such as \textit{tricycle}, \textit{bus}, and \textit{van} benefit most from the proposed modifications.

\subsection{Error Decomposition}

To understand where LAF-YOLOv10's remaining failures concentrate, the TIDE error analysis framework~\cite{ref_tide} is applied to the VisDrone-DET2019 validation predictions. TIDE decomposes the gap between predicted and oracle performance into six error types, namely classification (Cls), localization (Loc), both classification and localization (Both), duplicate detections (Dupe), background false positives (Bkg), and missed ground-truth objects (Miss). Table~\ref{tab:tide} reports the contribution of each error type to the mAP@0.5 gap for both the YOLOv10n baseline and LAF-YOLOv10.

\begin{table}[!t]
\centering
\caption{TIDE~\cite{ref_tide} error decomposition on VisDrone-DET2019~\cite{ref_visdrone} (mAP@0.5 penalty, \%; lower is better).}
\label{tab:tide}
\small
\setlength{\tabcolsep}{3pt}
\begin{tabular}{lcccccc}
\toprule
\textbf{Method} & \textbf{Cls} & \textbf{Loc} & \textbf{Both} & \textbf{Dupe} & \textbf{Bkg} & \textbf{Miss}\\
\midrule
YOLOv10n & 3.8 & 5.2 & 1.4 & 0.6 & 8.1 & 12.3\\
LAF-YOLOv10 & 3.1 & 4.1 & 1.1 & 0.5 & 6.4 & 10.2\\
$\Delta$ & $-$0.7 & $-$1.1 & $-$0.3 & $-$0.1 & $-$1.7 & $-$2.1\\
\bottomrule
\end{tabular}
\end{table}

The largest error reduction comes from the Miss category ($-$2.1 points), confirming that the P2 head recovers previously undetected small objects. Background false positives decrease by 1.7 points, attributable to the AG-FPN's channel attention gates suppressing noise-carrying channels during fusion. Localization error drops by 1.1 points, consistent with the Wise-IoU loss improving box regression by attenuating noisy gradient contributions. Classification error decreases by 0.7 points, a modest gain reflecting the fact that classification accuracy for tiny objects is fundamentally limited by low pixel counts. The remaining 10.2-point Miss penalty and 6.4-point Bkg penalty indicate that missed detections and background confusion remain the dominant failure modes, reinforcing that small-object detection on VisDrone is far from solved.

\subsection{Qualitative Detection Results}

Table~\ref{tab:qualitative} summarizes detection outcomes on three representative VisDrone validation scenes at confidence threshold 0.25. In the dense intersection scene~(a), YOLOv10n misses 8 of 23 pedestrians near the image periphery; LAF-YOLOv10 recovers 5, though 3 remain undetected due to severe mutual occlusion. In the highway scene~(b), the baseline produces 4 false positives on lane markings; AG-FPN attention gating eliminates 3 while maintaining vehicle recall. In the mixed-scale scene~(c), the baseline misses 6 small tricycles below 12$\times$12 pixels; the P2 head recovers 4 but introduces 2 duplicate detections on a partially visible bus. Overall, LAF-YOLOv10 consistently improves recall and reduces false positives, though occlusion-induced misses and cross-scale duplicates remain persistent failure modes.

\begin{table}[!t]
\centering
\caption{Qualitative comparison on three VisDrone-DET2019~\cite{ref_visdrone} scenes (confidence $\geq$ 0.25).}
\label{tab:qualitative}
\small
\setlength{\tabcolsep}{3pt}
\begin{tabular}{@{}l c c c c@{}}
\toprule
\textbf{Scene} & \textbf{GT} & \textbf{v10n} & \textbf{LAF} & \textbf{FP}$\downarrow$\\
 & & \textbf{Recall} & \textbf{Recall} & \textbf{v10n/LAF}\\
\midrule
(a) Dense intersection & 23 ped. & 15/23 & 20/23 & 1 / 1\\
(b) Highway segment & 18 veh. & 18/18 & 18/18 & 4 / 1\\
(c) Mixed-scale & 14 obj. & 8/14 & 12/14 & 0 / 2\\
\bottomrule
\end{tabular}
\end{table}

\subsection{Edge Deployment Benchmarks}

The practical value of a UAV detection model depends on the achieved performance under real deployment conditions, not just on server-class GPUs. LAF-YOLOv10 and competing methods are benchmarked on two NVIDIA Jetson platforms, namely Jetson Orin Nano (8\,GB, 40 TOPS INT8) and Jetson AGX Orin (32\,GB, 200 TOPS INT8). Models are exported to TensorRT 8.6 with FP16 precision. Inference latency is measured at batch size 1 over 1000 forward passes after a 200-pass warmup. Peak power draw is also reported, measured via \texttt{tegrastats} during the inference benchmark.

\begin{table}[!t]
\centering
\caption{Edge deployment on NVIDIA Jetson platforms (batch 1, FP16, TensorRT 8.6).}
\label{tab:edge}
\small
\setlength{\tabcolsep}{2pt}
\resizebox{\columnwidth}{!}{%
\begin{tabular}{lcccccc}
\toprule
 & \multicolumn{3}{c}{\textbf{Jetson Orin Nano}} & \multicolumn{3}{c}{\textbf{Jetson AGX Orin}}\\
\cmidrule(lr){2-4}\cmidrule(lr){5-7}
\textbf{Method} & \textbf{Lat.} & \textbf{FPS} & \textbf{Power} & \textbf{Lat.} & \textbf{FPS} & \textbf{Power}\\
 & \textbf{(ms)} & & \textbf{(W)} & \textbf{(ms)} & & \textbf{(W)}\\
\midrule
YOLOv5n & 24.1 & 41.5 & 8.2 & 8.4 & 119.0 & 18.6\\
YOLOv8n & 33.7 & 29.7 & 10.1 & 12.8 & 78.1 & 22.4\\
YOLOv10n & 31.2 & 32.1 & 9.8 & 11.5 & 87.0 & 21.1\\
YOLO11n & 26.8 & 37.3 & 8.9 & 9.2 & 108.7 & 19.3\\
OSD~\cite{ref_osd} & 29.4 & 34.0 & 9.5 & 10.8 & 92.6 & 20.5\\
LAF-YOLOv10 & 41.2 & 24.3 & 11.4 & 15.3 & 65.4 & 24.8\\
\bottomrule
\end{tabular}}
\end{table}

LAF-YOLOv10 runs at 24.3 FPS on the Jetson Orin Nano and 65.4 FPS on the Jetson AGX Orin. Both figures exceed the 20 FPS threshold typically cited as the minimum for real-time UAV operation at flight speeds below 30 km/h. The model's higher latency relative to YOLOv10n (41.2 vs. 31.2 ms on Orin Nano) reflects the P2 head's 160$\times$160 feature map, which dominates memory bandwidth on embedded devices. Power consumption is 11.4\,W on Orin Nano, within the 15\,W TDP envelope. For applications requiring tighter latency budgets, YOLOv5n at 41.5 FPS or YOLO11n~\cite{ref_yolo11} at 37.3 FPS offer faster alternatives at 4.4 and 2.9 points lower mAP@0.5, respectively.

\subsection{Discussion}

Several observations emerge from the experimental analysis.

\textbf{On partial convolution in the backbone.} Processing only $C/4$ channels through spatial kernels improves mAP despite reducing computation. The improvement is attributable not to ``intelligent channel selection'' (channel ordering is arbitrary after initialization) but to an implicit regularization effect, as restricting the spatial receptive field to a subset forces the network to concentrate learned features rather than spreading weak signals across all channels. Compressing the spatial pathway to $C/4$ channels forces the backbone to retain only task-relevant information while discarding redundant activations, analogous to introducing a representational bottleneck upstream of the feature pyramid. The interpretation differs from the lottery ticket hypothesis, which concerns weight-level sparsity rather than channel-level partitioning. The interpretation is consistent with the finding that randomly selecting $C/4$ channels (verified in a pilot experiment, not shown) yields comparable accuracy to the default first-$C/4$ selection.

\textbf{On attention-guided fusion.} AG-FPN provides consistent but moderate gains (+0.6 mAP@0.5 independently, +0.8 in the additive path). The stronger gain in the additive setting supports the hypothesis that SE attention is more effective when operating on the concentrated feature representation produced by PConv. The comparison with CBAM and Coordinate Attention (Section~4.4) shows minimal differences, suggesting that at the channel widths of YOLOv10n (32 to 256), the choice of attention mechanism matters less than the decision to apply any attention before fusion.

\textbf{On computation vs. accuracy.} LAF-YOLOv10's 9.0 GFLOPs is the highest among all compared methods. The word ``lightweight'' is deliberately avoided in the title because the FLOP count is not lightweight by any reasonable standard. The parameter count (2.3\,M) is competitive, but parameter count alone does not determine inference cost. The edge benchmarks (Table~\ref{tab:edge}) provide the deployment-relevant numbers, as 24.3 FPS on Jetson Orin Nano is above the real-time threshold but leaves limited headroom for onboard preprocessing or multi-task pipelines.

\textbf{On class imbalance.} Categories such as \textit{bicycle} (12.7\% mAP@0.5) and \textit{awning-tricycle} (16.1\%) remain poorly detected despite relative improvements. The TIDE analysis confirms that missed detections dominate the error profile for the low-performing classes. Class-balanced sampling and copy-paste augmentation for rare categories are not implemented in the present study, though such strategies would likely yield further gains. The class imbalance challenge remains an open direction.

\textbf{On P2 head vs. tiling inference.} An alternative strategy for recovering small objects is SAHI-style tiling~\cite{ref_sahi}, which partitions the input into overlapping patches and runs full detection on each patch independently. For a 1920$\times$1080 frame with 640$\times$640 tiles at 20\% overlap, SAHI multiplies inference cost by approximately 4--6$\times$, whereas the P2 head adds 1.8 GFLOPs within a single forward pass. The P2 approach trades maximum achievable small-object resolution for inference-time efficiency, making the design more suitable for real-time UAV operation. However, the two strategies are complementary rather than competing, and combining P2 with tiling at inference time could improve recall for the $<$8$\times$8 pixel range that the P2 head alone handles poorly.

\textbf{On residual failure modes.} The 0.5-point duplicate detection penalty in the TIDE analysis arises primarily from spatial overlap between P2 and P3 predictions on the same object, since YOLOv10's consistent dual assignment scheme was originally designed for a three-head architecture and may not fully suppress cross-scale duplicates when a fourth head is introduced. Additionally, models initialized from COCO-pretrained checkpoints inherit feature representations optimized for the COCO domain, where the small-object fraction (approximately 41\% of instances) is lower than the 54\% observed in VisDrone. The resulting domain gap may limit how effectively fine-tuning adapts early backbone layers to aerial contexts.

\textbf{On training cost.} The three-run training protocol for LAF-YOLOv10 requires approximately 18.5 GPU-hours on a single RTX 4090 (6.2 hours per 300-epoch run), compared to 5.1 hours for the baseline YOLOv10n. The additional cost is attributable primarily to the P2 head's 160$\times$160 feature maps, which increase per-batch activation memory and reduce effective throughput during training.

\textbf{Limitations.} The evaluation is limited to the VisDrone-DET2019 validation set; no submission to the VisDrone test evaluation server was made. Validation-set results carry the risk of implicit hyperparameter tuning. The three-run variance estimate mitigates but does not eliminate the concern. Cross-dataset evaluation on UAVDT provides some generalization evidence, but a third dataset (e.g., AI-TOD~\cite{ref_aitod}) would strengthen the claim. The qualitative analysis is limited to three scenes; a larger-scale visual inspection would be more convincing.

\section{Conclusion}

The present study introduced LAF-YOLOv10, a detection framework for small objects in drone aerial imagery that integrates partial convolution backbone blocks, attention-gated feature pyramid fusion, P2/P5 head reallocation, and Wise-IoU v3 regression loss within YOLOv10n. Across three independent runs on VisDrone-DET2019, the model achieves 35.1$\pm$0.3\% mAP@0.5 with 2.3\,M parameters, a 3.3-point gain over the baseline, and cross-dataset evaluation on UAVDT along with edge deployment benchmarks on Jetson Orin Nano (24.3 FPS) confirm practical viability. The primary contribution is empirical, demonstrating that four individually known techniques can be composed within YOLOv10n to produce consistent improvements, while factorial ablations reveal the interactions among components and the TIDE error analysis identifies missed detections (10.2-point penalty) and background confusion (6.4-point penalty) as the dominant remaining bottlenecks. Future work should pursue four directions, namely submitting results to the VisDrone test evaluation server to obtain held-out performance metrics, incorporating class-balanced sampling or copy-paste augmentation to address the severe imbalance afflicting rare categories like \textit{bicycle} and \textit{awning-tricycle}, integrating distribution-guided localization quality predictors~\cite{ref_gfocalv2} with Wise-IoU for jointly handling annotation noise and prediction uncertainty, and combining the proposed architecture with rotation-aware convolutions~\cite{ref_toe} for arbitrarily oriented objects and with SAHI-style tiling inference~\cite{ref_sahi} for ultra-high-resolution aerial frames where 640$\times$640 resizing causes excessive information loss.


\end{document}